\documentclass{article}
\usepackage{xcolor}
\usepackage{float}
\usepackage[final]{corl_2026}
\usepackage{amsmath}
\usepackage{graphicx}
\usepackage{colortbl}
\usepackage{wrapfig}
\usepackage{booktabs}
\usepackage{tabularx}
\usepackage{placeins}
\title{X-WBC: A Cross-Embodiment Foundation Model for Humanoid Whole-Body Control}
\author{
  Juntong Zhang$^{1,2,4}$ \qquad
  Chun Gu$^{1,3,4}$ \qquad
  Li Zhang$^{1,3,4,*}$\\[4pt]
  \small $^1$Shanghai Innovation Institute \qquad
  $^2$Tongji University\\
  \small $^3$Fudan University \qquad
  $^4$Logos Intelligence Lab\\[2pt]
  \small $^*$Corresponding author. \quad
  Project website: \url{https://logosintelligencelab.github.io/x-wbc/}
}

\hypersetup{pdftitle={X-WBC: A Cross-Embodiment Foundation Model for Humanoid Whole-Body Control},pdfauthor={Juntong Zhang, Chun Gu, Li Zhang},pdfkeywords={Whole-body control, Humanoid robots, Cross-embodiment learning}}
\begin{document}
\maketitle
\begingroup
\setlength{\intextsep}{2pt}
\vspace{-0.3in}
\begin{figure}[H]
  \centering
  \includegraphics[width=\textwidth,trim=0 0 0 12bp,clip]{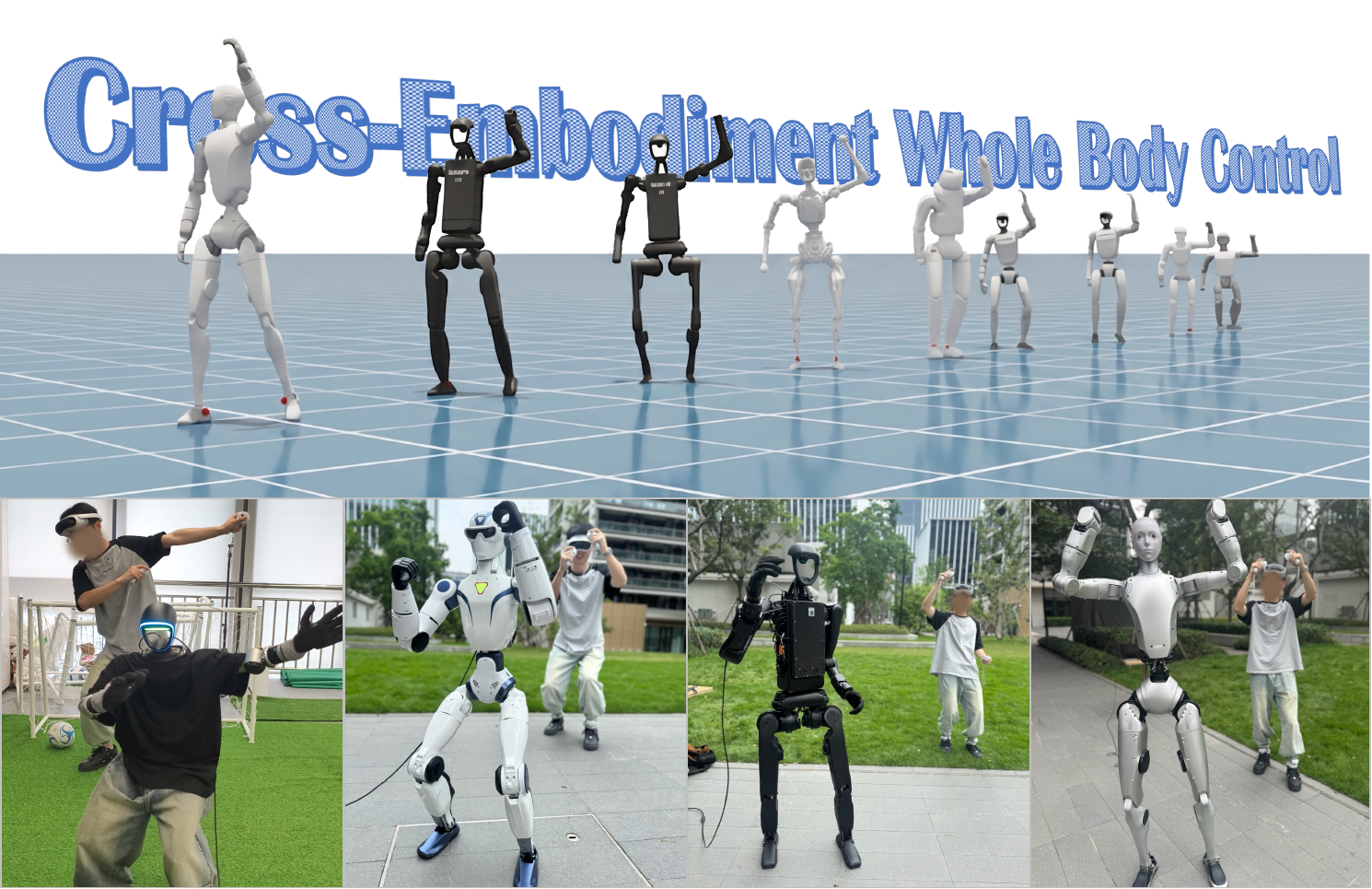}
  \caption{\textbf{X-WBC.} Multiple humanoid robots jointly train a shared whole-body controller in simulation. Shared modules learn reusable human-motion semantics, while robot-specific modules handle embodiment-dependent execution. Real-robot VR teleoperation demonstrates a common human-centered interface across substantially different bodies.}
  \label{fig:teaser}
\end{figure}
\endgroup

\begin{abstract}
Scaling humanoid whole-body control toward general-purpose deployment requires large human motion corpora and training experience shared across robot bodies.
Existing methods usually train one policy per robot, leaving motion experience isolated across embodiments.
We introduce X-WBC, a cross-embodiment foundation framework that separates relatively shared human motion semantics from embodiment-specific physical execution.
Human-centered command tokens align full human motion, robot reference motion, and sparse VR observations.
A causal Transformer learns reusable temporal structure from mixed multi-robot rollouts, while lightweight robot-specific modules map the shared representation to each robot's proprioception and action space.
Across nine simulated embodiments, external motions, and four real robots, experiments show that joint training improves tracking, the aligned representation supports consistent control across command sources, and the learned policy remains competitive beyond the training corpus.
These results support heterogeneous humanoids as joint data sources and establish cross-embodiment joint training as a practical route toward whole-body control foundation models.
\end{abstract}
\vspace{-0.8em}

\section{Introduction}

Does motor intelligence admit a representation that can transfer across bodies? Humans can learn motor skills by observing others despite differences in height, limb length, strength, and joint range. A martial arts student need not share the instructor's exact body proportions to understand the intent behind a kick, a turn, or a squat. We ask whether humanoid whole-body control (WBC) can learn a similar cross-embodiment motion representation. Such a representation would allow experience collected with one body to improve control learning for others instead of remaining tied to a single platform.

Recent humanoid WBC uses large human motion corpora and motion tracking to support teleoperation, skill learning, and autonomy~\citep{he2024h2o,he2024omnih2o,fu2024humanplus,liao2025beyondmimic,luo2025sonic}. These systems move WBC beyond hand-designed behaviors, but their training unit is still usually a single robot. Each platform retargets motions, trains policies, and tunes deployment around its own morphology, so the diversity in datasets such as AMASS, LAFAN1, and BONES-SEED~\citep{amass2019,lafan1,bonesstudio2026seed} becomes isolated robot-specific motion data. In parallel, robotics foundation models pursue cross-embodiment data sharing, using heterogeneous robot datasets to train shared backbones with embodiment-aware interfaces~\citep{brohan2023rt2,kim2024openvla,black2024pi0,bjorck2025gr00t,oneill2024openx,khazatsky2024droid,wu2025robomind,bu2025agibot,wang2024hpt,doshi2024crossformer,zheng2025xvla}. Whether this idea can benefit humanoid WBC remains largely unexplored.

Our central insight is that human motion data carries relatively embodiment-independent motion semantics, while the robot motion used to execute those semantics is embodiment-specific. Walking, kicking, squatting, and reaching share temporal structure and geometric intent across bodies; humanoids mainly differ in how these intents become executable motion through their kinematics, joint limits, actuation, and body layout. A cross-embodiment WBC system should therefore separate shared motion semantics from robot-specific motion generation. Retargeting one human motion across robots provides distinct realizations of the same intent, so robot diversity complements motion diversity as supervision for the shared backbone. If this separation holds, training signals from different robots can optimize a common motion model instead of remaining isolated in separate controllers.

We introduce \textbf{X-WBC}, a cross-embodiment WBC framework evaluated on nine humanoid robots. The framework casts training as multi-robot motion tracking, where rollouts from different humanoids jointly supervise a shared policy backbone. Its key design is a human-motion-centered command-token space that aligns full human motion, robot reference motion, and sparse VR motion as different views of the same underlying motion intent. A shared temporal Transformer learns the history-dependent motion structure needed for cross-robot understanding, while lightweight robot-specific modules adapt this representation to each robot's proprioception and action space. Training mixes transitions from all robots in one PPO~\citep{schulman2017ppo} batch, with command token alignment and adaptive motion sampling to stabilize joint optimization.

We validate X-WBC through simulation and deployment on multiple humanoid robots. Controlled comparisons examine whether sharing multi-robot experience improves tracking and whether the aligned command routes support consistent control. An external-motion evaluation tests the frozen policy beyond its training corpus, while token-space retrieval and real humanoid teleoperation examine the learned representation and its deployment on different robot bodies. Together, these results support multi-robot motion tracking as a practical foundation task for sharing whole-body motion experience across heterogeneous humanoids.

This paper makes three contributions: (1) a cross-embodiment WBC formulation that shifts the training unit from isolated per-robot policies to joint training across multiple humanoid robots. (2) A human-motion-centered policy architecture that separates shared motion semantics from robot-specific execution through aligned command tokens, a shared temporal backbone, and lightweight robot-specific modules. (3) Experiments showing the effects of joint training and core components, consistent multi-source control, competitive external-motion tracking, token-space retrieval, and real-world deployment.

\section{Related work}
\label{sec:relatedwork}

\textbf{Whole-body control.}
Recent humanoid whole-body control increasingly uses motion tracking as a low-level interface for teleoperation, skill learning, and autonomy.
H2O~\citep{he2024h2o}, OmniH2O~\citep{he2024omnih2o}, and HumanPlus~\citep{fu2024humanplus} build RGB and VR teleoperation, dexterous loco-manipulation, robot shadowing, and autonomous-skill collection on this interface, while BeyondMimic~\citep{liao2025beyondmimic} and SONIC~\citep{luo2025sonic} scale motion tracking toward composable behaviors and motion foundation tasks.
These systems, however, are still organized around a single embodiment: each platform retargets motion data and trains policies for its own morphology.
We study whether multiple humanoid bodies can instead share one WBC framework while retaining robot-specific execution.

\textbf{Cross-embodiment robot foundation models.}
Robot foundation models increasingly treat heterogeneous robot data as a scalable training mixture, with generalist policies such as RT-2~\citep{brohan2023rt2}, OpenVLA~\citep{kim2024openvla}, $\pi_0$~\citep{black2024pi0}, $\pi_{0.5}$~\citep{physicalintelligence2025pi05}, GR00T~\citep{bjorck2025gr00t}, and $\pi_{0.7}$~\citep{physicalintelligence2026pi07} learning reusable vision, language, and action mappings from large datasets such as Open X-Embodiment~\citep{oneill2024openx}, DROID~\citep{khazatsky2024droid}, RoboMIND~\citep{wu2025robomind}, and AgiBot~\citep{bu2025agibot}.
This mixture exposes heterogeneity in morphology, sensing, proprioception, action spaces, and data protocols, which HPT~\citep{wang2024hpt}, CrossFormer~\citep{doshi2024crossformer}, and X-VLA~\citep{zheng2025xvla} address with embodiment-aware stems, shared policies, or soft prompts.
These works motivate a trainable interface between shared representations and robot-specific inputs and outputs, but they mainly target manipulation and general vision-language-action settings.
We bring this perspective to closed-loop humanoid WBC, where shared motion representations must also respect balance, dynamics, and joint-space control.

\textbf{Human motion representation for humanoid control.}
Motion resources such as AMASS~\citep{amass2019}, LAFAN1~\citep{lafan1}, BONES-SEED~\citep{bonesstudio2026seed}, and SMPL~\citep{loper2015smpl} provide reusable motion semantics, but WBC must map them to robot morphology.
This problem has been studied for articulated characters~\citep{choi1999online,popovic1999motion,tak2005retargeting,villegas2018neural,aberman2020skeleton,hu2023pose} and in humanoid robotics through geometric matching, scaling, and inverse kinematics~\citep{penco2018robust,darvish2019whole,tang2024humanmimic,luo2023phc,protomotions2024,xie2025kungfubot,ze2025twist}.
GMR~\citep{araujo2025retargeting} further shows that retargeting quality strongly affects tracking and provides cleaner humanoid reference motions.
We use these robot motions not as isolated per-robot targets, but as one command route aligned with full human motion and sparse VR motion.

\FloatBarrier
\section{Method}
\label{sec:method}

\begin{figure*}[!htbp]
  \centering
  \includegraphics[width=\textwidth,trim=20bp 1bp 0 1bp,clip]{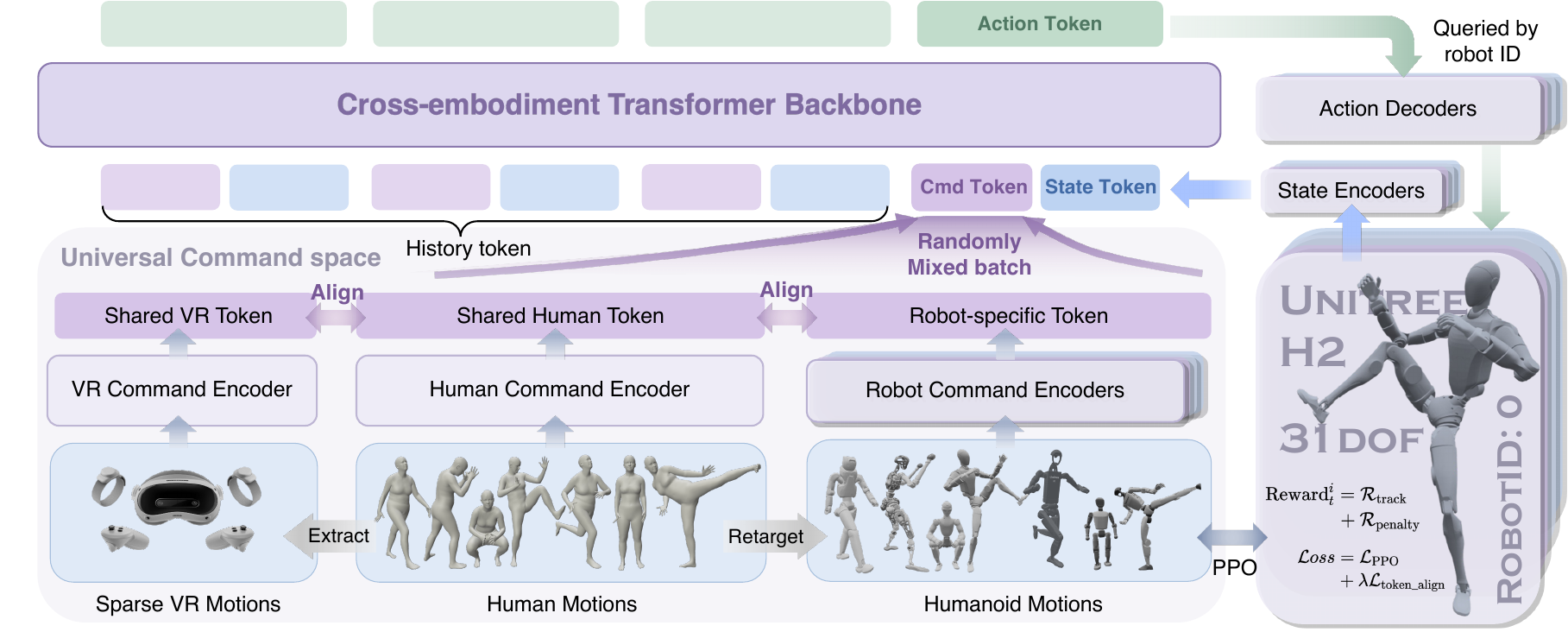}
  \caption{\textbf{Overview of X-WBC.} Full human motion, robot motion commands, and sparse VR commands are encoded into a shared motion-token space. A shared Transformer backbone integrates the mixed token batch with robot proprioception and history, while robot-specific modules route each sample to the proper embodiment interface and decode executable joint actions.}
  \label{fig:method_overview}
\end{figure*}

As shown in Figure~\ref{fig:method_overview}, the framework treats cross-embodiment whole-body control as human-motion-centered representation learning rather than isolated policy training for each humanoid morphology. A shared motion Transformer learns relatively embodiment-independent motion semantics, while robot-specific modules realize these semantics under each robot's proprioception, dynamics, and action space. We first formulate the multi-robot tracking problem (Section~\ref{sec:method_formulation}), then describe command encoding (Section~\ref{sec:method_command_encoding}), the shared backbone and robot-specific modules (Section~\ref{sec:method_backbone}), and the joint training procedure (Section~\ref{sec:method_training}).

\subsection{Cross-embodiment whole-body motion tracking}
\label{sec:method_formulation}

We separate relatively embodiment-independent human motion semantics from embodiment-specific robot motion generation. To make this separation learnable, we cast training as a multi-robot motion-tracking problem, where related motions executed by different humanoids provide supervision for a shared motion Transformer rather than remaining isolated in robot-specific controllers. Formally, each robot $i$ defines an MDP $\mathcal{M}_i=(\mathcal{S}_i,\mathcal{A}_i,\mathcal{T}_i,\mathcal{R},\gamma)$, where the state space $\mathcal{S}_i$, action space $\mathcal{A}_i$, and transition dynamics $\mathcal{T}_i$ are embodiment-specific, while the reward family $\mathcal{R}$ and discount factor $\gamma$ are shared across robots. We optimize the shared tracking objective over rollouts from all robots with PPO~\citep{schulman2017ppo}.

\textbf{Observations.}
We use an asymmetric actor-critic observation design. The actor observation $s_t^i$ consists of robot proprioception $s_t^{p,i}$ and one motion command $s_t^c$. The proprioception $s_t^{p,i}=(q_t^i,\dot{q}_t^i,\omega_t^i,g_t^i,a_{t-1}^i)$ contains joint positions $q_t^i$, joint velocities $\dot{q}_t^i$, base angular velocity $\omega_t^i$, gravity projected in the body frame $g_t^i$, and the previous action $a_{t-1}^i$. The command $s_t^c$ is selected from robot motion, full human motion, and sparse VR motion. Position, orientation, and velocity commands are expressed in a root-centered or robot-centric frame by default. The critic receives the actor observation plus privileged terms for value estimation, including robot body-link positions and orientations in the body frame and base linear velocity.

\textbf{Actions.}
Each policy output is interpreted in the action space of the active robot. For robot $i$, the policy outputs valid joint actions that are converted to torques by a low-level PD controller. Since robots have different observation and action dimensions, these vectors are padded to the maximum dimensions in the robot set during training so that transitions from multiple robots can be mixed in one batch. At execution time, the robot-specific decoder keeps only the valid action dimensions of the corresponding robot.

\textbf{Rewards.}
The reward uses the same tracking semantics across robots. For robot $i$, the reward can be summarized as $r_t^i=\mathcal{R}_{\mathrm{track}}+\mathcal{P}$, where $r_t^i$ is the scalar reward at time $t$, $\mathcal{R}_{\mathrm{track}}$ groups tracking terms for root orientation, body-link pose, and body velocity, and $\mathcal{P}$ groups penalties for action-rate changes, joint-limit violations, and undesired contacts. The full reward list is provided in Appendix.

\subsection{Human-centered command encoding}
\label{sec:method_command_encoding}

We align motion commands from different sources into a human-centered motion-token space. Human motion provides a common description before robot-specific retargeting, while the shared policy aggregates experience from different robot bodies. Alignment makes the command routes compatible with that shared policy. We use three command encoder routes: full human motion provides shared motion semantics, robot motion provides reference motions in robot body spaces, and sparse VR provides a deployable human control interface.

\textbf{Human motion command encoder.}
This shared encoder maps full-body human motion to a human motion token. Its input comes from motion capture and provides a unified humanoid motion description. Because this token is not tied to any robot joint definition, it serves as the semantic anchor for command alignment while preserving human motion intent and temporal structure.

\textbf{Robot motion command encoder.}
This encoder maps reference motions in robot body spaces to robot motion tokens. These references usually come from human-to-robot retargeting, but can also come from robot recordings, generated motions, or other robot motion sources. Since robots differ in body definitions, joint orderings, and executable constraints, this encoder is robot-specific.

\textbf{Sparse VR command encoder.}
This shared encoder maps sparse observations of human motion to a sparse VR token. During training, we extract five human keypoints, the head and four limb endpoints, from full human motion. During deployment, the same signal format can be obtained from VR devices in real time. Five tracked points provide a low-bandwidth control interface without requiring full motion capture or per-motion retargeting at deployment. Dense robot references remain necessary for training supervision, and this interface does not remove robot-specific calibration or control setup.

Across all routes, target motion is represented with body positions, orientations, and velocities rather than only robot joint angles, providing the spatial and velocity information needed for precise tracking. Before entering the encoders, these quantities are normalized in a root-centered or robot-centered frame, so the tokens focus on relative motion structure and remain robust to global-coordinate shifts and different operator body sizes. During training, command token alignment encourages human, robot, and sparse VR tokens that describe the same underlying motion to be used consistently by the shared policy backbone.

\subsection{Shared policy backbone and robot-specific modules}
\label{sec:method_backbone}

Many humanoid motion-tracking policies use MLPs to predict actions from the current state and reference motion. This single-step mapping is effective for local tracking on one robot, but cross-embodiment control must learn shared motion structure across multiple robots and command forms. Historical context is therefore important for resolving motion phase, velocity trends, and temporal dependencies. We use an attention-based temporal backbone over observation-token histories, allowing the shared policy to learn reusable temporal motion patterns from rollouts across all robots.

The shared policy backbone is implemented as a causal Transformer~\citep{vaswani2017transformer}. For robot $i$, each control step first combines the selected command token $z_{c,t}$ and the robot-specific proprioception token $z_{p,t}^i$ into the current observation token. The Transformer aggregates current and past observation tokens within a causal history window of length $L$ and outputs an action token for action decoding:
\begin{equation}
h_t^i =
B_{\theta}\left(
\{[z_{c,\tau}; z_{p,\tau}^i]\}_{\tau=t-L+1}^{t}
\right),
\label{eq:backbone}
\end{equation}
where $B_{\theta}$ is the policy backbone shared by all robots, and $h_t^i$ is the action token output at the current observation-token position. During online rollout, a KV cache stores the keys and values of previous tokens, so each control step only processes the new observation token and updates the cache instead of recomputing the full history window. This substantially reduces computation for real-time control while preserving multi-frame context.

Proprioception encoding and action decoding are robot-specific because humanoids differ in state dimensions, joint ordering, actuator limits, and action spaces. Like the robot motion command encoder, both are lightweight robot-specific modules selected by robot ID: the former maps padded proprioceptive observations to a fixed-dimensional proprioception token, and the latter maps the action token $h_t^i$ to valid joint actions for robot $i$. This division lets the shared Transformer learn reusable cross-embodiment motion semantics, while the robot-specific modules translate those semantics into executable joint actions for each robot.

\subsection{Training}
\label{sec:method_training}

\textbf{Multi-robot joint training.}
We use Isaac Lab~\citep{mittal2025isaac} to construct a multi-robot training environment. Multiple robots perform rollouts within the same training framework, while each robot keeps its own state, action, contact, and termination logic. Transitions from all robots are collected after each simulation step and mixed into the same PPO~\citep{schulman2017ppo} batch, which jointly updates the shared modules and the active robot-specific modules.

\textbf{Optimization.}
For each motion clip, the policy encodes robot motion, full human motion, and sparse VR motion into three command tokens. During rollout, one token is sampled with a 1:1:1 ratio and used as the Transformer backbone input throughout the episode. The training objective combines the PPO tracking objective with command token alignment:
\begin{equation}
\mathcal{L}
=
\mathcal{L}_{\mathrm{PPO}}
+
\lambda_{\mathrm{align}}\mathcal{L}_{\mathrm{align}}.
\label{eq:loss}
\end{equation}
Here $\mathcal{L}_{\mathrm{PPO}}$ includes the clipped policy loss, value regression, and entropy regularization. $\mathcal{L}_{\mathrm{align}}$ computes pairwise MSE among the three normalized command tokens from the same motion clip before they enter the backbone. Full training hyperparameters are listed in Appendix.

\textbf{Termination.}
Episode termination removes clear tracking failures and improves training efficiency. We terminate an episode when root vertical position, root orientation, or end-effector vertical position error exceeds its threshold. These events also provide failure statistics for adaptive motion sampling. The training thresholds are listed in Appendix~\ref{app:termination}; the common external-evaluation rule is specified in Section~\ref{sec:exp_setup}.

\textbf{Adaptive motion sampling.}
Following failure-rate-based sampling used in motion tracking~\citep{liao2025beyondmimic,luo2025sonic}, we increase the probability of motion segments that frequently trigger early termination. This focuses training on difficult clips while retaining coverage of the full motion set. Because persistent failures can also indicate bad retargeting, incorrect contacts, or physically infeasible segments, we remove consistently low-success segments from the training pool. The sampling and hard-removal parameters are listed in Appendix.

\textbf{Domain randomization.}
To improve sim-to-real robustness and disturbance recovery, we use moderate domain randomization. We randomize contact parameters, default joint positions, and base center-of-mass offsets, and periodically apply root velocity perturbations. All random variables are sampled from uniform distributions, with ranges listed in Appendix.

\section{Experiments}
\label{sec:experiments}
We evaluate X-WBC from four complementary perspectives: policy design and joint training, transfer to external motions, the structure of the learned command space, and real-robot deployment. Together, these experiments test whether multi-robot training produces a reusable control backbone while preserving robot-specific execution.

\subsection{Experimental setup}
\label{sec:exp_setup}
We train in Isaac Lab on nine humanoid embodiments: Unitree G1 variants, H1, H1-2, R1, and H2, together with Fourier GR3, Booster T1, and Adam Lite. Training uses about 200 hours of BONES-SEED motion~\citep{bonesstudio2026seed}, resampled at 50 Hz. We adapt the GMR retargeting pipeline~\citep{araujo2025retargeting} through smoothing and robot-specific parameters. All policies use PPO, with 1024 parallel environments per robot on 8 NVIDIA H100 GPUs. The final model is trained for approximately two days, runs at 50 Hz, and uses a 32-step history. Joint training mixes the active robots' rollouts into a shared update batch. Appendix~\ref{app:method_details} provides implementation details.

\textbf{Evaluation data.}
Table~\ref{tab:main_sim_results} evaluates Unitree G1 and H2 on the BONES-SEED motions used for training. These results measure in-distribution architecture and fitting behavior, rather than generalization to an independent motion corpus. For external evaluation, 100STYLE~\citep{mason2022local} contains one actor performing 100 locomotion styles. Our fixed grid pairs each style with eight movement categories, giving 800 clips and 133.02 minutes at 50 Hz. Table~\ref{tab:external_results} evaluates the complete grid with all policy weights frozen.

\textbf{Protocol and metrics.}
All systems use the same G1 reference trajectories, clip windows, and method-independent scoring. A clip terminates when pelvis-height error exceeds 0.25 m, the vertical projected-gravity components differ by more than 0.8, or any wrist or ankle height error exceeds 0.25 m. We use one deterministic rollout per clip (seed 42, flat terrain, no evaluation randomization), with completion taking priority at the final frame. Success rate (SR) and mean clip completion measure rollout survival; MPKPE and body velocity error are averaged over all active steps, including failed partial rollouts. Early failures censor these errors, so tracking errors are interpreted together with SR and completion.

\subsection{Architecture, joint training, and ablations}
\label{sec:exp_results}
Table~\ref{tab:main_sim_results} separates policy architecture and joint training, command-source consistency, and component removal on the BONES-SEED training motions.
\begin{table*}[t]
  \caption{\textbf{Motion tracking on BONES-SEED training motions.} Transformer is the default backbone; X-WBC (MLP) replaces the shared Transformer with an MLP, while Single robot uses the same Transformer architecture as X-WBC but is trained per robot. Src. denotes the evaluation command source. $\mathrm{SR}$ is success rate (\%), and $E_{\mathrm{mpkpe}}$ (m), $E_{\mathrm{vel}}$ (m/s), $E_{\mathrm{root}}^p$ (m), and $E_{\mathrm{root}}^R$ (rad) measure mean per-keypoint position, body velocity, root position, and root rotation errors over all evaluated rollouts.}
  \label{tab:main_sim_results}
  \centering
  \begingroup
  \scriptsize
  \definecolor{g1blue}{rgb}{0.90,0.955,1.00}
  \definecolor{h2purple}{rgb}{0.94,0.90,0.98}
  \definecolor{sectiongray}{rgb}{0.93,0.93,0.93}
  \setlength{\tabcolsep}{2.0pt}
  \renewcommand{\arraystretch}{1.08}
  \resizebox{\textwidth}{!}{%
  \begin{tabular}{@{}l c >{\columncolor{g1blue}}c >{\columncolor{g1blue}}c >{\columncolor{g1blue}}c >{\columncolor{g1blue}}c >{\columncolor{g1blue}}c >{\columncolor{h2purple}}c >{\columncolor{h2purple}}c >{\columncolor{h2purple}}c >{\columncolor{h2purple}}c >{\columncolor{h2purple}[\tabcolsep][0pt]}c@{}}
    \hline
    \multicolumn{2}{c}{\textbf{Setting}} & \multicolumn{5}{>{\columncolor{g1blue}}c}{\textbf{Unitree G1 (29DoF)}} & \multicolumn{5}{>{\columncolor{h2purple}[\tabcolsep][0pt]}c}{\textbf{Unitree H2}} \\
    \hline
    \textbf{Model} & \textbf{Src.}
    & $\mathbf{SR}(\%)\uparrow$ & $\mathbf{E}_{\mathbf{mpkpe}}\downarrow$ & $\mathbf{E}_{\mathbf{vel}}\downarrow$ & $\mathbf{E}_{\mathbf{root}}^{p}\downarrow$ & $\mathbf{E}_{\mathbf{root}}^{R}\downarrow$
    & $\mathbf{SR}(\%)\uparrow$ & $\mathbf{E}_{\mathbf{mpkpe}}\downarrow$ & $\mathbf{E}_{\mathbf{vel}}\downarrow$ & $\mathbf{E}_{\mathbf{root}}^{p}\downarrow$ & $\mathbf{E}_{\mathbf{root}}^{R}\downarrow$ \\
    \hline
    \rowcolor{sectiongray}\multicolumn{12}{@{}l@{}}{\textbf{\textit{Policy architecture and joint training}}} \\
    X-WBC & VR
      & \textbf{98.60} & 0.0412 & \textbf{0.2120} & 0.9936 & \textbf{0.4354} & \textbf{93.22} & \textbf{0.0629} & \textbf{0.4172} & 1.4716 & \textbf{0.4535} \\
    X-WBC (MLP) & VR
      & 97.70 & \textbf{0.0411} & 0.2272 & \textbf{0.9433} & 0.4465 & 91.33 & 0.0652 & 0.4542 & \textbf{1.4316} & 0.4833 \\
    Single robot & VR
      & 97.69 & 0.0412 & 0.2195 & 0.9756 & 0.4431 & 92.22 & 0.0698 & 0.4417 & 1.7709 & 0.4978 \\
    \hline
    \rowcolor{sectiongray}\multicolumn{12}{@{}l@{}}{\textbf{\textit{Command source consistency}}} \\
    X-WBC & Robot
      & 98.21 & 0.0409 & 0.2164 & 1.1591 & \textbf{0.3554} & 92.08 & 0.0634 & 0.4274 & 1.6865 & \textbf{0.3758} \\
    X-WBC & Human
      & \textbf{98.66} & \textbf{0.0397} & \textbf{0.2098} & \textbf{0.9562} & 0.4053 & \textbf{93.02} & \textbf{0.0610} & \textbf{0.4124} & \textbf{1.4069} & 0.4270 \\
    \hline
    \rowcolor{sectiongray}\multicolumn{12}{@{}l@{}}{\textbf{\textit{Ablations}}} \\
    w/o human and robot routes & VR
      & 95.34 & 0.0492 & 0.2494 & 1.1032 & 0.5093 & 85.71 & 0.0722 & 0.5016 & 1.6970 & 0.5299 \\
    w/o human route & VR
      & 95.17 & 0.0500 & 0.2508 & 1.1437 & 0.5125 & 83.84 & 0.0763 & 0.5244 & 1.7784 & 0.5342 \\
    w/o robot route & VR
      & 95.33 & 0.0503 & 0.2461 & 1.1760 & 0.5046 & 85.21 & 0.0757 & 0.5106 & 1.7719 & 0.5229 \\
    w/o alignment loss & VR
      & 95.41 & 0.0496 & 0.2469 & 1.1128 & 0.5029 & 85.68 & 0.0735 & 0.5100 & 1.6512 & 0.5297 \\
    w/o robot-specific modules & VR
      & 96.54 & 0.0474 & 0.2241 & 1.0611 & 0.4793 & 88.65 & 0.0724 & 0.4575 & 1.5091 & 0.4875 \\
    \hline
  \end{tabular}}
  \endgroup
\end{table*}

\begin{table*}[t]
\caption{\textbf{External-motion evaluation on the complete 800-clip 100STYLE grid (G1).} X-WBC uses its default sparse-VR command source; external systems retain their native inputs and low-level controllers. Errors include all active steps before termination; position is in m and velocity in m/s.}
\label{tab:external_results}
\centering
\begingroup
\fontsize{6pt}{7pt}\selectfont
\definecolor{g1blue}{rgb}{0.90,0.955,1.00}
\setlength{\tabcolsep}{3pt}
\renewcommand{\arraystretch}{1.08}
\begin{tabularx}{\textwidth}{@{}>{\raggedright\arraybackslash}p{0.15\textwidth}>{\raggedright\arraybackslash}p{0.14\textwidth}>{\centering\arraybackslash\columncolor{g1blue}}X>{\centering\arraybackslash\columncolor{g1blue}}X>{\centering\arraybackslash\columncolor{g1blue}}X>{\centering\arraybackslash\columncolor{g1blue}[\tabcolsep][0pt]}X@{}}
\hline
\textbf{Model} & \textbf{Src.} & $\mathbf{SR}(\%)\uparrow$ & $\mathbf{Comp.}(\%)\uparrow$ & $\mathbf{E}_{\mathbf{mpkpe}}\downarrow$ & $\mathbf{E}_{\mathbf{vel}}\downarrow$ \\
\hline
X-WBC & VR5 & 91.13 & 91.66 & 0.0649 & 0.5400 \\
SONIC & SMPL & 93.00 & 93.35 & 0.0687 & 0.3419 \\
TWIST & SMPL--GMR & 75.62 & 78.18 & 0.0565 & 0.5261 \\
\hline
\end{tabularx}
\endgroup
\end{table*}

\FloatBarrier
\begin{figure}[!t]
  \centering
  \input{figures/embodiment_coverage}
  \input{figures/real_world_deployment}
\end{figure}

\textbf{Architecture and joint training.}
On G1, X-WBC reaches 98.60\% SR, compared with 97.70\% for X-WBC (MLP) and 97.69\% for Single robot. On H2, the corresponding rates are 93.22\%, 91.33\%, and 92.22\%. The full model also gives the lowest body-velocity error on both robots. Similar G1 position errors but better success and velocity tracking indicate improved closed-loop stability, while the larger H2 gap shows the value of shared supervision on a harder embodiment. Transfer beyond the training motions is examined separately below.

\textbf{Command sources.}
For the same full model, the VR, robot-motion, and full-human routes achieve similar training-motion success on both G1 and H2, showing that the aligned encoders provide compatible inputs to the shared backbone. Human commands are slightly stronger on several body metrics, whereas robot commands give lower root-rotation error because they are already in the target reference space. Sparse VR remains close despite using only five keypoints. Robot commands here are an inference-time input to the jointly trained model, not a separately trained robot-only policy.

\textbf{Component removal.}
Removing either or both dense command routes, the alignment loss, or the robot-specific modules lowers training-motion SR on both robots. The largest H2 reductions occur when a dense command route is removed. The relatively strong result without robot-specific modules suggests that proprioception itself carries useful embodiment information. Together, these results support the complete design on the evaluated training distribution.
\subsection{External-motion evaluation}
Table~\ref{tab:external_results} compares X-WBC using its default sparse-VR interface with released SONIC~\citep{luo2025sonic} and TWIST~\citep{ze2025twist} under common scoring. The external systems retain their native inputs and low-level controllers: SMPL for SONIC and SMPL-derived motion through GMR for TWIST. The evaluated releases and checkpoints available in our pipeline cover Unitree G1, so we conduct the matched external comparison on that platform. Neither baseline reports 100STYLE as a training source.

On the complete 100STYLE grid, X-WBC with VR5 commands completes 729 clips (91.13\% SR), SONIC 744 (93.00\%), and TWIST 605 (75.62\%). SONIC has lower velocity error, whereas TWIST has lower MPKPE but substantially lower completion. Because stepwise errors stop accumulating after termination, TWIST's lower MPKPE should be read alongside its 15.51-point SR deficit to X-WBC. X-WBC therefore trades a small position-error gap for much broader rollout survival on the complete grid, while remaining close to SONIC in success.

\subsection{Embodiment coverage and command-token semantics}
\label{sec:exp_token_semantics}
Figure~\ref{fig:embodiment_coverage} reports the final shared Transformer's training-motion performance across all nine robots. Success and MPKPE vary across platforms, but the shared policy achieves broad coverage across markedly different bodies. Their rankings need not coincide: success measures whether a rollout avoids termination, whereas MPKPE measures tracking accuracy during active steps. Reporting both distinguishes robust rollout survival from precise motion tracking.

The right panel evaluates whether the learned tokens retain motion identity across input sources and robot embodiments. Cross-source retrieval reaches 86.0\% Recall@1 for the same motion phase and 94.2\% for the same motion. Cross-robot retrieval reaches 61.3\% and 76.1\%, above the corresponding random baselines. The consistently higher same-motion recall indicates that motion identity is more stable than exact temporal phase. Lower cross-robot than cross-source recall further suggests that embodiment-specific execution information remains in the tokens. These results show that the aligned token space captures shared motion structure while retaining the embodiment information needed for execution.

\subsection{Real-world deployment}
\label{sec:exp_real}
We deploy X-WBC on Unitree G1, R1, H1-2, and H2 using the same policy architecture and sparse-VR interface. Figure~\ref{fig:real_world_deployment} shows upper-body pose control and lower-body motions on robots with different sizes and joint counts. Across platforms, the operator uses the same command format, while robot-specific branches map it to each robot's proprioception and action space. These demonstrations show that the same human-centered command interface and shared backbone can support robots with substantially different scales and kinematic structures. Quantitative cross-method hardware comparisons and cross-manufacturer deployment remain directions for broader evaluation.

\section{Conclusion}
\label{sec:conclusion}
X-WBC reframes humanoid whole-body control as a cross-embodiment learning problem. Its shared motion Transformer learns temporal structure from mixed multi-robot experience, aligned human-centered command tokens connect dense motion references with deployable sparse inputs, and lightweight robot-specific modules translate the shared representation into executable actions. Across simulation, external-motion evaluation, token-space analysis, and deployment on four real humanoids, the results show that a single jointly trained backbone can improve tracking, remain competitive beyond the training motions, and provide one control interface across bodies with substantially different sizes and kinematics. These findings establish multi-robot motion tracking as a practical route toward foundation models that accumulate, rather than isolate, whole-body control experience across humanoid platforms.

\section{Limitations}
\label{sec:limitations}
The current study focuses on humanoid robots, so the extent to which the learned motion representation transfers to more diverse morphologies, body proportions, and joint layouts remains open. Although the shared backbone accumulates experience across robots, X-WBC still uses robot-specific command and proprioception encoders, action decoders, control logic, and per-robot motion retargeting to preserve each platform's executable action space. These interfaces remain part of the deployment pipeline, and their reduction should be evaluated without weakening control fidelity. Real-world deployment has been validated on four Unitree robots and a limited set of upper- and lower-body behaviors. Broader hardware evidence should include other manufacturers, longer task sequences, and more varied contacts and terrains. Future work will study how transfer scales across broader robot sets, reduce robot-specific preparation, and systematically evaluate robustness to sensing dropouts, communication delay, state-estimation noise, and persistent disturbances. A further direction may be zero-shot X-WBC, in which a policy trained across known embodiments controls an unseen humanoid without training a new robot-specific module. Achieving this goal may require morphology-conditioned interfaces that derive sensing and action mappings from a robot description while preserving stable closed-loop execution.

\acknowledgments{This work was supported in part by the New Generation Artificial Intelligence--National Science and Technology Major Project (2025ZD0123004), the National Natural Science Foundation of China (Grant No.~62376060), and the Ningbo grant (2025Z038). We thank Professor Xin Ruan of the College of Civil Engineering at Tongji University for helpful discussions and support.}

\clearpage
\bibliography{main}
\appendix
\setcounter{table}{0}
\renewcommand{\thetable}{A\arabic{table}}
\renewcommand{\theHtable}{appendix.A\arabic{table}}
\section{Appendix}
\label{app:method_details}
\setlength{\textfloatsep}{8pt plus 2pt minus 2pt}
\setlength{\floatsep}{8pt plus 2pt minus 2pt}
\setlength{\intextsep}{8pt plus 2pt minus 2pt}
\setlength{\abovecaptionskip}{4pt}
\setlength{\belowcaptionskip}{2pt}

This appendix reports supplementary implementation details for the X-WBC model, including reward and termination settings, training configuration, sampling and domain randomization, and network details.

\subsection{Reward}
\label{app:reward_termination}

The same reward family is used for all robots so that different embodiments provide comparable tracking supervision to the shared policy backbone. The tracking terms measure root position and orientation, body pose, and body velocity in the command frame, while regularization terms discourage abrupt actions, joint-limit violations, and contacts on non-allowed links. Table~\ref{tab:app_reward_terms} lists the reward terms.

\begin{table}[H]
\caption{\textbf{Reward design.} Superscript $g$ denotes the goal or command reference, superscript $p$ denotes the current robot state, $\mathcal{B}$ is the tracked body set, and $\mathcal{C}_{\mathrm{allow}}$ is the set of bodies allowed to make contact.}
\label{tab:app_reward_terms}
\centering
\small
\setlength{\tabcolsep}{2pt}
\renewcommand{\arraystretch}{1.0}
\begin{tabularx}{\textwidth}{@{}p{0.21\textwidth}>{\raggedright\arraybackslash}X>{\centering\arraybackslash}p{0.10\textwidth}@{}}
\hline
\textbf{Reward term} & \textbf{Equation} & \textbf{Weight} \\
\hline
\multicolumn{3}{@{}l}{\textit{Tracking rewards}} \\
\quad Root position &
$r^{\mathrm{root}}_{\mathrm{pos}}(t)=\exp\!\big(-\|\mathbf{p}^{g}_{t,\mathrm{root}}-\mathbf{p}^{p}_{t,\mathrm{root}}\|_2^2/0.3^2\big)$ &
0.5 \\
\quad Root orientation &
$r^{\mathrm{root}}_{\mathrm{ori}}(t)=\exp\!\big(-\|\mathbf{o}^{g}_{t,\mathrm{root}}-\mathbf{o}^{p}_{t,\mathrm{root}}\|_2^2/0.4^2\big)$ &
0.5 \\
\quad Body pos (rel.) &
$r^{\mathrm{body}}_{\mathrm{pos}}(t)=\exp\!\big(-\frac{1}{|\mathcal{B}|}\sum_{b\in\mathcal{B}}\|\mathbf{p}^{g,\mathrm{rel}}_{t,b}-\mathbf{p}^{p,\mathrm{rel}}_{t,b}\|_2^2/0.3^2\big)$ &
1.0 \\
\quad Body ori (rel.) &
$r^{\mathrm{body}}_{\mathrm{ori}}(t)=\exp\!\big(-\frac{1}{|\mathcal{B}|}\sum_{b\in\mathcal{B}}\|\mathbf{o}^{g,\mathrm{rel}}_{t,b}-\mathbf{o}^{p,\mathrm{rel}}_{t,b}\|_2^2/0.4^2\big)$ &
1.0 \\
\quad Body lin. vel &
$r^{\mathrm{body}}_{\mathrm{lin}}(t)=\exp\!\big(-\frac{1}{|\mathcal{B}|}\sum_{b\in\mathcal{B}}\|\mathbf{v}^{g}_{t,b}-\mathbf{v}^{p}_{t,b}\|_2^2/1.0^2\big)$ &
1.0 \\
\quad Body ang. vel &
$r^{\mathrm{body}}_{\mathrm{ang}}(t)=\exp\!\big(-\frac{1}{|\mathcal{B}|}\sum_{b\in\mathcal{B}}\|\boldsymbol{\omega}^{g}_{t,b}-\boldsymbol{\omega}^{p}_{t,b}\|_2^2/3.14^2\big)$ &
1.0 \\
\multicolumn{3}{@{}l}{\textit{Penalty terms}} \\
\quad Action rate &
$r_{\mathrm{act}}(t)=\|\mathbf{a}_t-\mathbf{a}_{t-1}\|_2^2$ &
-0.1 \\
\quad Joint limit &
$r_{\mathrm{jlim}}(t)=\sum_j \mathbf{1}[q_{t,j}\notin[q^{\min}_{j},q^{\max}_{j}]]$ &
-10.0 \\
\quad Undesired contacts &
$r_{\mathrm{contact}}(t)=\sum_{k\notin\mathcal{C}_{\mathrm{allow}}}\mathbf{1}[\|\mathbf{f}_{k}\|>1.0\mathrm{N}]$ &
-0.1 \\
\hline
\end{tabularx}
\end{table}

\subsection{Training termination}
\label{app:termination}

Episode termination removes clear tracking failures from rollout collection and provides failure statistics for adaptive sampling. We use three robot-independent training checks: root height error, root orientation error, and vertical end-effector error relative to the command reference. Table~\ref{tab:app_termination} lists these settings.

\begin{table}[H]
\caption{\textbf{Training termination settings.} Superscripts follow Table~\ref{tab:app_reward_terms}; $\mathcal{E}$ is the checked end-effector set, and $d_R$ computes angular distance between orientations.}
\label{tab:app_termination}
\centering
\small
\setlength{\tabcolsep}{2pt}
\renewcommand{\arraystretch}{1.0}
\begin{tabularx}{\textwidth}{@{}p{0.32\textwidth}>{\raggedright\arraybackslash}X>{\centering\arraybackslash}p{0.14\textwidth}@{}}
\hline
\textbf{Condition} & \textbf{Error measure} & \textbf{Threshold} \\
\hline
Root vertical position error &
$|(\mathbf{p}^{p}_{t,\mathrm{root}}-\mathbf{p}^{g}_{t,\mathrm{root}})_z|$ &
0.25 m \\
Root orientation error &
$d_R(\mathbf{o}^{p}_{t,\mathrm{root}},\mathbf{o}^{g}_{t,\mathrm{root}})$ &
0.8 rad \\
End-effector vertical position error &
$\max_{e\in\mathcal{E}} |(\mathbf{p}^{p}_{t,e}-\mathbf{p}^{g}_{t,e})_z|$ &
0.25 m \\
\hline
\end{tabularx}
\end{table}

\subsection{Training hyperparameters}
\label{app:training_hyperparameters}

Optimizer-level PPO parameters used in the main experiments are listed in Table~\ref{tab:app_training_setup}. The normalized pairwise token-alignment loss uses $\lambda_{\mathrm{align}}=1.0$, with equal weight for all three token pairs. Architecture, padding, command routing, sampling, and randomization settings are reported in the following appendix sections.

\begin{table}[H]
\caption{\textbf{PPO hyperparameters.} Values are the default settings used for the main multi-robot Transformer policy unless otherwise stated.}
\label{tab:app_training_setup}
\centering
\small
\setlength{\tabcolsep}{2pt}
\renewcommand{\arraystretch}{1.0}
\begin{tabularx}{\textwidth}{@{}>{\raggedright\arraybackslash}X>{\centering\arraybackslash}p{0.14\textwidth}@{\hspace{6pt}\vrule width 0.35pt\hspace{6pt}}>{\raggedright\arraybackslash}X>{\centering\arraybackslash}p{0.14\textwidth}@{}}
\hline
\textbf{Item} & \textbf{Value} & \textbf{Item} & \textbf{Value} \\
\hline
Parallel environments per robot & 1024 & Rollout steps per environment & 24 \\
Command route sampling ratio & 1:1:1 & Learning epochs & 5 \\
Number of minibatches & 4 & Discount factor $\gamma$ & 0.99 \\
GAE $\lambda$ & 0.95 & PPO clip ratio & 0.2 \\
Desired KL & 0.01 & Entropy coefficient $\lambda_e$ & 0.005 \\
Value loss coefficient $\lambda_v$ & 1.0 & Shared PPO learning rate & 1e-3 \\
Adaptive learning rate range & [1e-5, 1e-3] & Max gradient norm & 1.0 \\
Initial action noise std & 1.0 & Actor std min clamp & 1e-6 \\
\hline
\end{tabularx}
\end{table}

\subsection{Adaptive motion sampling}
\label{app:adaptive_sampling}

Adaptive motion sampling is used to avoid spending most updates on already solved motion segments while still retaining broad coverage of the motion dataset. We first maintain a motion-level working set to control memory use. Within the active working set, each motion is divided into temporal bins, and the bin-level sampling probability is increased for segments that frequently trigger early termination. When drawing a new working set, bin probabilities are aggregated to the motion level and normalized by the number of bins in each motion, preventing long clips from dominating only because they contain more bins. A small uniform mixture is retained for exploration. Segments that remain persistently infeasible are removed from the active pool by hard removal, which mainly filters severe retargeting artifacts and physically implausible snippets.

\begin{table}[H]
\caption{\textbf{Adaptive sampling.} The values summarize the failure-rate-based sampling schedule used for the main policy.}
\label{tab:app_adaptive_sampling}
\centering
\small
\setlength{\tabcolsep}{2pt}
\renewcommand{\arraystretch}{1.0}
\begin{tabularx}{\textwidth}{@{}p{0.32\textwidth}>{\centering\arraybackslash}p{0.22\textwidth}>{\raggedright\arraybackslash}X@{}}
\hline
\textbf{Item} & \textbf{Value} & \textbf{Role} \\
\hline
Motion working-set size & 1024 motions & Limits the active motion pool \\
Working-set refresh interval & $1\times10^9$ transitions & Makes reloads effectively rare \\
Bin duration & 1.0 s & Defines the failure-statistics window \\
Pre-failure sampling window & 4.0 s & Samples before recent failures \\
Initial bin probability & smoothed prior (1.0) & Starts close to uniform \\
Failure-rate cap & 50$\times$ mean & Limits failure-driven emphasis \\
Uniform mixture & 0.1 & Preserves broad motion coverage \\
Maximum bin probability & $50/N_{\mathrm{valid}}$ & Caps each valid bin \\
Minimum evaluation count & 96 attempts & Requires enough samples \\
Failure rate for removal & 0.98 & Excludes consistently failed bins \\
Hard-bin minimum cap hits & 32 & Requires repeated cap hits \\
Global hard-bin warmup & 500 PPO iterations & Delays global exclusion \\
\hline
\end{tabularx}
\end{table}

\subsection{Domain randomization}
\label{app:domain_randomization}

Domain randomization is applied during training to improve robustness to model mismatch, external perturbations, and noisy motion commands.

\begin{table}[H]
\caption{\textbf{Domain randomization.} $\mathcal{U}[\cdot]$ denotes a uniform distribution.}
\label{tab:app_domain_randomization}
\centering
\small
\setlength{\tabcolsep}{2pt}
\renewcommand{\arraystretch}{1.0}
\begin{tabularx}{\textwidth}{@{}p{0.34\textwidth}>{\raggedright\arraybackslash}X@{}}
\hline
\textbf{Domain randomization} & \textbf{Sampling distribution} \\
\hline
\multicolumn{2}{@{}l}{\textit{Physical parameters}} \\
\quad Static friction coefficients & $\mu_s \sim \mathcal{U}[0.3,1.6]$ \\
\quad Dynamic friction coefficients & $\mu_d \sim \mathcal{U}[0.3,1.2]$ \\
\quad Restitution coefficient & $e \sim \mathcal{U}[0,0.5]$ \\
\quad Default joint positions & $\mathbf{q}_0 \leftarrow \mathbf{q}_0 + \mathcal{U}[-0.01,0.01]$ \\
\quad Base COM offset $(x,y,z)$ & $\Delta x \sim \mathcal{U}[-0.075,0.075],\ \Delta y,\Delta z \sim \mathcal{U}[-0.1,0.1]$ \\
\multicolumn{2}{@{}l}{\textit{Root velocity perturbations}} \\
\quad Root linear velocity $(x,y,z)$ & $v_x,v_y \sim \mathcal{U}[-0.5,0.5],\ v_z \sim \mathcal{U}[-0.2,0.2]$ \\
\quad Push duration & $\Delta t \sim \mathcal{U}[1,3]\,\mathrm{s}$ \\
\quad Root angular velocity & $\omega_{\mathrm{roll}},\omega_{\mathrm{pitch}} \sim \mathcal{U}[-0.52,0.52],\ \omega_{\mathrm{yaw}} \sim \mathcal{U}[-0.78,0.78]$ \\
\hline
\end{tabularx}
\end{table}

\begin{table}[H]
\caption{\textbf{Robot embodiment summary.} Action dimensions are reported before shared padding; heights and masses are approximate public specifications.}
\label{tab:app_padding_summary}
\centering
\small
\setlength{\tabcolsep}{3pt}
\renewcommand{\arraystretch}{1.0}
\begin{tabularx}{\textwidth}{@{}>{\raggedright\arraybackslash}Xccc@{\hspace{8pt}}>{\raggedright\arraybackslash}Xccc@{}}
\hline
\textbf{Robot} & \textbf{Dim.} & \textbf{Height} & \textbf{Mass} &
\textbf{Robot} & \textbf{Dim.} & \textbf{Height} & \textbf{Mass} \\
\hline
Unitree G1-29DOF & 29 & 132 cm & 35 kg & Unitree H2 & 27 & 182 cm & 70 kg \\
Unitree G1-23DOF & 23 & 132 cm & 35 kg & Booster T1 & 23 & 118 cm & 30 kg \\
Unitree H1 & 19 & 180 cm & 47 kg & Fourier GR3 & 32 & 165 cm & 71 kg \\
Unitree H1-2 & 27 & 178 cm & 70 kg & PND Adam Lite & 31 & 167 cm & 60 kg \\
Unitree R1 & 28 & 123 cm & 29 kg & & & & \\
\hline
\end{tabularx}
\end{table}

\subsection{Network details}
\label{app:network_padding}

The default policy maps each command source and the robot proprioception into compact tokens, then applies a Transformer backbone for history-dependent control. The full-human and VR command encoders are shared across robots, while the robot-command encoder, proprioception encoder, and output heads are routed by robot domain.

\begin{table}[H]
\caption{\textbf{Input dimensions.} For motion commands, each body or keypoint contributes 12 values: position 3, 6D orientation 6, and linear velocity 3.}
\label{tab:app_input_dims}
\centering
\small
\setlength{\tabcolsep}{2pt}
\renewcommand{\arraystretch}{1.0}
\begin{tabularx}{\textwidth}{@{}p{0.34\textwidth}>{\centering\arraybackslash}p{0.12\textwidth}>{\raggedright\arraybackslash}X@{}}
\hline
\textbf{Input item} & \textbf{Dim.} & \textbf{Source} \\
\hline
Full-human motion command & 264 & 22 joints $\times$ 12 \\
Sparse-VR motion command & 60 & 5 keypoints $\times$ 12 \\
Robot motion command & 168 & 14 robot bodies $\times$ 12 \\
Proprioception & 102 & padded robot state features \\
Privileged critic observation & 399 & asymmetric critic input \\
Action & 32 & padded joint action \\
\hline
\end{tabularx}
\end{table}

\begin{table}[H]
\caption{\textbf{Network architecture.} The default Transformer uses separate actor and critic backbones with causal sliding-window attention. The MLP baseline keeps the same encoders and changes the backbone paths.}
\label{tab:app_network_details}
\centering
\small
\setlength{\tabcolsep}{2pt}
\renewcommand{\arraystretch}{1.0}
\begin{tabularx}{\textwidth}{@{}p{0.31\textwidth}p{0.15\textwidth}>{\raggedright\arraybackslash}X@{}}
\hline
\textbf{Module} & \textbf{Type} & \textbf{Size or configuration} \\
\hline
\multicolumn{3}{@{}l}{\textbf{Transformer-backbone}} \\
\hline
Command/state encoders (4) & MLP & each [256, 256, 64] \\
Actor fusion / critic embedding & Linear & [128 $\rightarrow$ 224] / [399 $\rightarrow$ 224] \\
Actor / critic backbones & Transformer & separate [224] backbones; depth 3, heads 8, head dim 32, window 32 \\
Actor / critic output heads & Linear & [224 $\rightarrow$ 32] / [224 $\rightarrow$ 1] \\
\hline
\multicolumn{3}{@{}l}{\textbf{MLP-backbone}} \\
\hline
Actor fusion / critic embedding & Linear & [128 $\rightarrow$ 256] / [399 $\rightarrow$ 256] \\
Actor / critic backbones & MLP & separate [1024, 768, 384] backbones \\
Actor / critic output heads & Linear & [384 $\rightarrow$ 32] / [384 $\rightarrow$ 1] \\
\hline
\end{tabularx}
\end{table}

\end{document}